\documentclass[conference,a4paper,10pt]{IEEEtran}

\IEEEoverridecommandlockouts
\usepackage{cite}
\usepackage{amsmath,amssymb,amsfonts}
\usepackage{algorithmic}
\usepackage{graphicx}
\usepackage{textcomp}
\usepackage{xcolor}
\usepackage{makecell}
\usepackage{threeparttable}
\usepackage{float}
\usepackage{stfloats}
\usepackage{array}
\usepackage{marvosym}
\usepackage{geometry}

\def\BibTeX{{\rm B\kern-.05em{\sc i\kern-.025em b}\kern-.08em
    T\kern-.1667em\lower.7ex\hbox{E}\kern-.125emX}}
\newenvironment{shrinkeq}[1]
{ \bgroup
	\addtolength\abovedisplayshortskip{#1}
	\addtolength\abovedisplayskip{#1}
	\addtolength\belowdisplayshortskip{#1}
	\addtolength\belowdisplayskip{#1}}
{\egroup\ignorespacesafterend}
\begin{document}
\title{Subject-independent Human Pose Image Construction with Commodity Wi-Fi\\
}

\author{\IEEEauthorblockN{Shuang Zhou, Lingchao Guo, \Letter{Zhaoming Lu}, Xiangming Wen, Wei Zheng, Yiming Wang}
	\IEEEauthorblockA{Beijing Laboratory of Advanced Information Networks,\\ Beijing Key Laboratory of Network System Architecture and Convergence,\\ School of Information and Communication Engineering,\\ Beijing University of Posts and Telecommunications, Beijing, China.\\ E-mail: {\{zhoushuang, rita\_guo, lzy0372, xiangmw, zhengweius, wym\_1997\}}@bupt.edu.cn}}
\maketitle

\begin{abstract}
Recently, commodity Wi-Fi devices have been shown to be able to construct human pose images, i.e., human skeletons, as fine-grained as cameras. Existing papers achieve good results when constructing the images of subjects who are in the prior training samples. However, the performance drops when it comes to new subjects, i.e., the subjects who are not in the training samples. This paper focuses on solving the subject-generalization problem in human pose image construction. To this end, we define the subject as the domain. Then we design a Domain-Independent Neural Network (DINN) to extract subject-independent features and convert them into fine-grained human pose images. We also propose a novel training method to train the DINN and it has no re-training overhead comparing with the domain-adversarial approach. We build a prototype system and experimental results demonstrate that our system can construct fine-grained human pose images of new subjects with commodity Wi-Fi in both the visible and through-wall scenarios, which shows the effectiveness and the subject-generalization ability of our model.
\end{abstract}

\begin{IEEEkeywords}
human pose image construction, DINN, subject-generalization, subject-independent, commodity Wi-Fi
\end{IEEEkeywords}
\vspace{-0.2cm}

\section{Introduction}
Human poses can provide useful information for tasks such as human-computer interaction, medical care, and autonomous driving\cite{yang20183d}, so human pose image construction is indispensable research in the human sensing area. Recently, due to the prevalence of Wi-Fi infrastructures, privacy-protection and through-wall abilities of Wi-Fi signals, human pose image construction with commodity Wi-Fi has attracted extensive attention in academia and industry. 

Because of being reflected, scattered and diffracted by objects or human bodies in the ambient environment, Wi-Fi signals propagate from the transmitter to the receiver through multipath. Thus, the received superposition signals carry information reflecting the characteristics of the propagation space, including the human pose information. Channel State Information (CSI) obtained from commodity Wi-Fi Network Interface Cards (NICs) by open-source tool\cite{halperin2011tool} mainly represents the received superposition signal, which provides the possibility for human pose image construction with Wi-Fi.

Past papers\cite{guo2019signal,wang2019person,guo2020healthcare} have achieved fine-grained 2D human pose construction with commodity Wi-Fi through deep learning methods. However, when it comes to new subjects, these models have poor performance, i.e., they do not have subject-generalization ability. \cite{jiang2020towards} constructs the 3D human poses with commodity Wi-Fi through regressing the 3D positions of human joints directly and explores the generalization ability of the model they proposed. However, the subject performs activities on a fixed spot in their work. So, it remains to be verified whether the model can adapt to subjects who move in the entire perceptual space.

Therefore, to achieve fine-grained and subject-independent human pose image construction with commodity Wi-Fi is still challenging: the differences among subjects such as height, weight, gender and clothing, affect Wi-Fi signals in different ways, which will sharply decline the generalization ability of the models. 

\textit{If we define human subjects as domains, the above differences can be considered as domain differences}. Then the domain-adversarial networks can be adopted to solve the above problem. To recognize human activities in different environments, \cite{jiang2018towards} proposes an EI framework, which uses the domain-adversarial training approach to extract the environment-independent features of human activities, and then they utilize several types of signals to demonstrate the effectiveness of their framework. \cite{zhao2017learning} proposes a conditional adversarial architecture which retains all information relevant to the predictive task through discarding the information specific to domains. 

Nevertheless, the domain-adversarial training approach will add significant overhead in the model re-training process, which is not applicable in real life. It is necessary for domain-adversarial training methods to feed source domain and unlabeled target domain data into the network for training at the same time\cite{ganin2016domain}. The network builds feature mappings between the source and the target domains by doing so.

Given these analyses, we focus on achieving fine-grained and subject-independent human pose image construction with commodity Wi-Fi and avoiding overhead in the model re-training process. The main contributions are listed as follows:

$1.$ We design the Domain-Independent Neural Network (DINN), a deep learning network which can extract domain-independent features and construct fine-grained human pose images. Specifically, the DINN is composed of feature extractor, generator, and domain discriminator. The feature extractor is used to extract low-dimensional features related to human poses. The generator is used to convert the features into human pose images and the domain discriminator is designed to distinguish which domain the features belong to. \textit{In this paper, we define the subject as the domain.} Actually, the domain discriminator is used to distinguish which subject the features belong to. Through training, the DINN can extract subject-independent features and convert them into fine-grained human pose images.

$2.$ We propose a novel training method to train the DINN. It includes two training stages in which only the source domain data is necessary for training. Therefore, our training method has no re-training overhead comparing with the domain-adversarial approach.

$3.$ We build a prototype system to construct fine-grained human pose images of new subjects with commodity Wi-Fi. Comprehensive experiments are conducted in both the visible and through-wall scenarios. Compared with the state-of-the-art method proposed in\cite{guo2020healthcare}, the average performance increases 37\% and 35.7\% on the strict match in the visible and the through-wall scenario, respectively. 65.99\% (50.27\%) and 100\% (99.82\%) human pose images of the new subject constructed by our system strictly and loosely match the ground truth in the visible (through-wall) scenario.	

The rest of this paper is organized as follows. Preliminaries and motivation are discussed in Section \uppercase\expandafter{\romannumeral2} and Section \uppercase\expandafter{\romannumeral3}. Section \uppercase\expandafter{\romannumeral4} elaborates the DINN and the training method. In Section \uppercase\expandafter{\romannumeral5}, we conduct a series of experiments in both the visible and through-wall scenarios to evaluate the performance of our system. Finally, we conclude the paper in Section \uppercase\expandafter{\romannumeral6}.
\vspace{-0.10cm}
\begin{figure*}
	\centering
	\setlength{\abovecaptionskip}{0.cm}
	\includegraphics[width=0.85\linewidth]{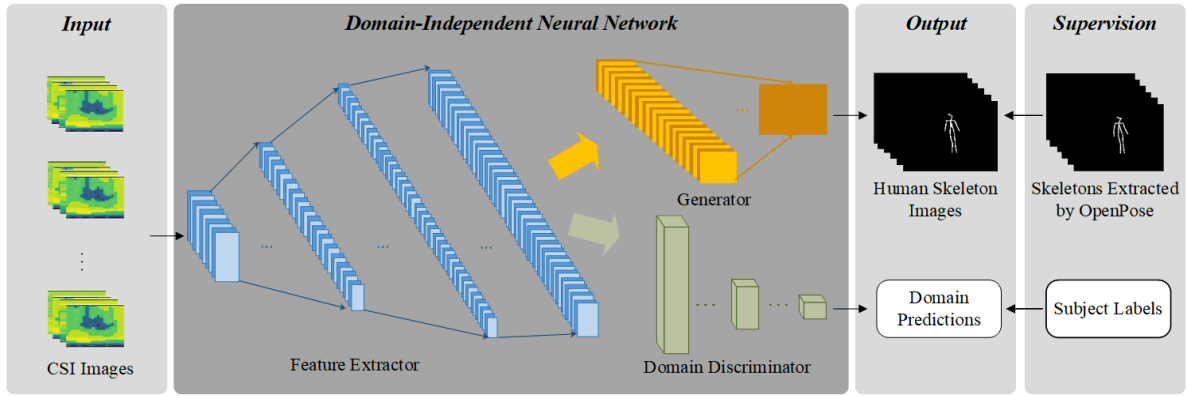}
	\caption{Overall framework of the DINN}
	\label{DINN}
	\vspace{-1.5em}
\end{figure*}
\section{Preliminaries}
\subsection{Channel State Information}
CSI is the equispaced samples of Channel Frequency Response (CFR), and it can be presented as follow:

\vspace{-1ex}
\begin{shrinkeq}{-1ex}
\begin{small}
	\begin{equation}
	\label{CSI}
	\setlength{\abovedisplayskip}{0pt}
	\setlength{\belowdisplayskip}{0pt}
	H(f,t)=\sum_{i=0}^{L}H_i(f,t)=\sum_{i=0}^{L}\alpha_i(t)e^{-j2\pi{f\tau_i(t)}},
	\end{equation}
\end{small}\end{shrinkeq}where $L$ represents the number of propagation paths and $\alpha{(t)}$ is the attenuation and $\tau{(t)}$ is the propagation delay.  

According to \cite{wang2015understanding}, CSI can be divided into dynamic paths and static paths based on whether their lengths are changed because of human body parts moving. Hence, equation (\ref{CSI}) can be expressed as follow:

\vspace{-1ex}
\begin{shrinkeq}{-1ex}
\begin{small}
	\begin{equation}
	\setlength{\abovedisplayskip}{0pt}
	\setlength{\belowdisplayskip}{0pt}
	\begin{split}
	H(f,t)&=H_s(f,t)+H_d(f,t)\\
	      &=H_s(f,t)+\sum_{i\in{L_d}}\alpha_i(t)e^{-j2\pi{f\tau_i(t)}},
	\end{split}
	\end{equation}
\end{small}\end{shrinkeq}where $H_s(f,t)$ represents the static component which consists of the direct path and paths reflected by static objects in the environment. $H_d(f,t)$ represents the dynamic component which is composed of reflection paths from the moving body, i.e., dynamic paths. And $L_d$ is the set of dynamic paths.
\vspace{-0.15cm}
\subsection{Data Collection and Processing}
When different subjects are moving in the sensing space, we collect CSI samples from commercial Wi-Fi NICs while we exploit a synchronous camera to record his (or her) video footage. 

We process the amplitude and phase of CSI samples and combine them to obtain the dynamic component. After signal processing, we segment dynamic components to form the CSI images according to the synchronous time information. Besides, we get human skeleton images of recorded video footage through OpenPose\cite{cao2017realtime}. For more data processing details, please refer to the previous work of our team\cite{guo2019signal}. 
\vspace{-0.15cm}
\section{Motivation}
Existing 2D human pose image construction models do not have subject-generalization ability. Because these models are fed the dynamic components for human pose image construction. However, differences among subjects will lead to different dynamic paths even if the same pose is performed. In this situation, different dynamic paths, i.e., different dynamic components, will lead to different distributions of extracted features. 

Consequently, the extracted features are not only relevant to human poses but also subjects. The subject-related features will sharply decline the subject-generalization ability of the models and result in poor performance when these models are used to new subjects. 

Therefore, to achieve subject-generalization in human pose image construction, we design the DINN to extract subject-independent features and convert them into fine-grained human pose images. Besides, in order to avoid re-training overhead and enhance the practicability of the DINN, we propose a novel training method with two stages. Finally, we build a prototype system to construct fine-grained human pose images of new subjects with commodity Wi-Fi, which shows the effectiveness and the subject-generalization ability of our model. 
\vspace{-0.15cm}
\section{Methodology}
In this paper, we design the DINN with a novel training method to extract subject-independent features with no re-training overhead and convert them into fine-grained human pose images.

We use CSI images as the input of our model and define them as $X$. The output is the constructed human skeleton images and predicted domains, which are defined as $y$ and $d$, respectively. We use human skeleton images extracted from synchronous video footage by OpenPose as annotations for constructed images, which are defined as $Y$. One-hot probability vectors of subjects are used as subject labels which are the supervision for predicted domains and defined as $D$. Note that skeleton images are used as annotations because they are more fault-tolerant than key points and we use grey-scale skeleton images for simplifying. The overall framework is shown in Figure \ref{DINN}. The details of our model and the training method will be elaborated in the rest of this section. 
\vspace{-0.15cm}
\subsection{Feature Extractor}
The feature extractor is used to transform the CSI images into low-dimensional features \begin{math}Z\end{math}. It mainly consists of several convolutional layers, which are widely used to distill low-dimensional features of human motions\cite{wang2018spatial}. Using \begin{math}\theta_f\end{math} to denote the set of parameters in the feature extractor, given the input data $X$, we can get the low-dimensional features as:

\vspace{-1ex}
\begin{shrinkeq}{-1ex}
\begin{small}
\begin{equation}
\setlength{\abovedisplayskip}{0pt}
\setlength{\belowdisplayskip}{0pt}
Z=G_f(X,\theta_f),
\end{equation}
\end{small}\end{shrinkeq}where \begin{math}G_f\end{math} denotes the feature extractor network.
\vspace{-0.15cm}
\subsection{Generator}
The generator is designed to convert the learned features, i.e., $Z$, into human pose images. It mainly consists of several resize convolution layers\cite{aitken2017checkerboard} instead of traditional transposed convolution layers. As they can eliminate the Checkerboard Artifacts\cite{odena2016deconvolution} and increase the resolution of constructed images. Let $\theta_g$ be the set of parameters in the generator and we can obtain the human pose images:

\vspace{-1ex}
\begin{shrinkeq}{-1ex}
\begin{small}
	\begin{equation}
	\setlength{\abovedisplayskip}{0pt}
	\setlength{\belowdisplayskip}{0pt}
	y=G_g(Z,\theta_g),
	\end{equation}
\end{small}\end{shrinkeq}where \begin{math}G_g\end{math}  denotes the generator network. Further, the difference between the constructed pose images and the ground truth is calculated by the cross-entropy loss function, which denotes as follow:

\vspace{-1ex}
\begin{shrinkeq}{-1ex}
\begin{small}
	\begin{equation}
	\setlength{\abovedisplayskip}{0pt}
	\setlength{\belowdisplayskip}{0pt}
	\begin{split}
	\label{equation:L_g}
	L_g(y,Y)=&\frac{1}{M}\sum_{m=0}^{M}\sum_{n=0}^{N}Y_{mn}\log{\frac{1}{y_{mn}}}\\
	         &+(1-Y_{mn})\log{\frac{1}{1-y_{mn}}},
	\end{split} 
	\end{equation}
\end{small}\end{shrinkeq}where $M$ is the number of human skeleton images in the mini batch, and $N$ is the number of pixels on each image. Therefore, our proposed model needs to optimize the feature extractor and the generator networks by minimizing $L_g$ in order to obtain fine-grained human pose images.
\vspace{-0.15cm}
\subsection{Domain Discriminator}
The domain discriminator leverages the learned features $Z$ as input, and aims at distinguishing which domain the features belong to. It consists of several fully connected layers, which are widely used for classification tasks\cite{krizhevsky2012imagenet}. Let $\theta_d$ be the set of parameters in the domain discriminator, we can obtain the predicted domains:

\vspace{-1ex}
\begin{shrinkeq}{-1ex}
\begin{small}
	\begin{equation}
	\setlength{\abovedisplayskip}{0pt}
	\setlength{\belowdisplayskip}{0pt}
	d=G_d(Z,\theta_d),
	\end{equation}
\end{small}\end{shrinkeq}where $G_d$ denotes the domain discriminator network. Further, we design the cross-entropy loss function to calculate the difference between domain predictions and truthful domain labels, which denotes as follow:

\vspace{-1ex}
\begin{shrinkeq}{-1ex}
\begin{small}
	\begin{equation}
	\setlength{\abovedisplayskip}{0pt}
	\setlength{\belowdisplayskip}{0pt}
	\begin{split}
	\label{equation:L_d}
	L_d(d,D)=&\frac{1}{M}\sum_{m=0}^{M}\sum_{k=0}^{K}D_{mk}\log{\frac{1}{d_{mk}}}\\
	         &+(1-D_{mk})\log{\frac{1}{1-d_{mk}}},
	\end{split} 
	\end{equation}
\end{small}\end{shrinkeq}where $K$ denotes the number of domains. Thus, the domain discriminator can get the maximum domain discrimination performance by minimizing $L_d$.
\vspace{-0.15cm}
\subsection{Training Method}
We propose a novel training method to train the DINN, which can be divided into two stages: pre-training and adversarial training.

During the pre-training stage, we leverage two optimizers to minimize $L_g$ and $L_d$, respectively. Thus, the feature extractor will extract features relevant to human poses while the domain discriminator will obtain maximum domain discrimination performance. It means that the domain discriminator can recognize the domain-related features as much as possible. The optimization equations are expressed as follows:

\vspace{-0.5ex}
\begin{shrinkeq}{-1ex}
\begin{small}
	\begin{equation}
	\setlength{\abovedisplayskip}{0pt}
	\setlength{\belowdisplayskip}{0pt}
	\label{equation:argminL_g}
    (\hat{\theta}_f,\hat{\theta}_g)=\mathop{argmin}_{\theta_f,\theta_g}\:L_g(\theta_f,\theta_g),
    \end{equation}
\end{small}
\end{shrinkeq}
\begin{shrinkeq}{-0.5ex}
\begin{small}
	\begin{equation}
	\setlength{\abovedisplayskip}{-0pt}
	\setlength{\belowdisplayskip}{0pt}
	\label{equation:argminL_d}   
	(\hat{\theta}_d)=\mathop{argmin}_{\theta_d}\:L_d(\hat{\theta}_f,\theta_d),
    \end{equation}
\end{small}\end{shrinkeq}where $\hat{\theta}_f,\hat{\theta}_g$ and $\hat{\theta}_d$ are saddle points of the model parameters that we are seeking \cite{ganin2016domain}. And they can be obtained by the following gradient updates:

\vspace{-1ex}
\begin{shrinkeq}{-1ex}
\begin{small}
	\begin{equation}
	\setlength{\abovedisplayskip}{-0pt}
	\setlength{\belowdisplayskip}{0pt} 
	\label{f_update}
    \theta_f\;\leftarrow\;\theta_f-\mu_1\frac{\partial L_g}{\partial \theta_f},
    \end{equation}
\end{small}
\end{shrinkeq}
\begin{shrinkeq}{-0.5ex}
\begin{small}
	\begin{equation}
	\setlength{\abovedisplayskip}{-0pt}
	\setlength{\belowdisplayskip}{0pt} 
	\label{g_update}
    \theta_g\;\leftarrow\;\theta_g-\mu_1\frac{\partial L_g}{\partial \theta_g},
    \end{equation}
\end{small}
\end{shrinkeq}
\begin{shrinkeq}{-1ex}
\begin{small}
	\begin{equation}
	\setlength{\abovedisplayskip}{-0pt}
	\setlength{\belowdisplayskip}{0pt}
	\label{d_update}
    \theta_d\;\leftarrow\;\theta_d-\mu_2\frac{\partial L_d}{\partial \theta_d},
    \end{equation}
\end{small}\end{shrinkeq}where $\mu_1,\mu_2$ are the learning rates.

However, getting the maximum domain discrimination performance exactly contradicts with our goal. Thus, based on equation (\ref{equation:L_g}) and (\ref{equation:L_d}), we define the joint loss function $L$ as follow:

\vspace{-1ex}
\begin{shrinkeq}{-1ex}
\begin{small}
	\begin{equation}
	\setlength{\abovedisplayskip}{-0pt}
	\setlength{\belowdisplayskip}{0pt}
	\label{loss_L}
    L=L_g-\lambda{L_d},
    \end{equation}
\end{small}\end{shrinkeq}where $\lambda$ is the adversarial parameter, which is a positive number and used to achieve the trade-off between the feature extractor and the domain discriminator in the learning process.

At the adversarial training stage, we use one optimizer to minimize the loss $L$, which needs to minimize $L_g$ and maximize $L_d$. Hence, the performance of human pose image construction is improved while the feature extractor tries its best to cheat the domain discriminator. On the other hand, we also use another optimizer to minimize the loss $L_d$, which will maximize the loss $L$, conversely. So, the domain discriminator also aims at identifying domain-related features as much as possible. Compared with the optimization equation (\ref{equation:argminL_g}) and (\ref{equation:argminL_d}) in the pre-training stage, we only modify (\ref{equation:argminL_g}) here, as follow:

\vspace{-0.5ex}
\begin{shrinkeq}{-1ex}
	\begin{small}
		\begin{equation}
		\setlength{\abovedisplayskip}{-0pt}
		\setlength{\belowdisplayskip}{0pt}
		(\hat{\theta}_f,\hat{\theta}_g)=\mathop{argmin}_{\theta_f,\theta_g}\:L(\theta_f,\theta_g,\hat{\theta}_d).
		\end{equation}
	\end{small}
    \vspace{-0.5em}
\end{shrinkeq}

As for gradient updates, compared with the equation (\ref{f_update}), (\ref{g_update}) and (\ref{d_update}) in the pre-training stage, only (\ref{f_update}) is modified here, as follow:

\vspace{-1ex}
\begin{shrinkeq}{-1ex}
\begin{small}
	\begin{equation}
	\setlength{\abovedisplayskip}{-0pt}
	\setlength{\belowdisplayskip}{0pt}
    \theta_f\;\leftarrow\;\theta_f-\mu_1(\frac{\partial L_g}{\partial \theta_f}-\lambda\frac{\partial L_d}{\partial \theta_f}).
\end{equation}
\end{small}
\vspace{-0.5em}
\end{shrinkeq}
%
%

Through this minimax game, the domain discrimination accuracy declines sharply. It means that the domain discriminator with superior performance cannot identify which domain the learned features belong to. Intuitively, the learned features are domain-independent. As mentioned above, the subject represents the domain. Therefore, the learned features are actually subject-independent and they are then converted into fine-grained human pose images in the generator. Note that only several source domain data are used for training in the above method. Consequently, our proposed model can significantly avoid re-training overhead.
\vspace{-0.1cm}

\section{Experiments and Performance}
\vspace{-0.1cm}
\subsection{Setup}
In our experiments, we employ three transceivers equipped with Intel $5300$ NICs (one transmitter and two receivers). The two pairs of transceivers are placed vertically to improve the spatial resolution. $1$ transmitting antenna are used at the transmitter to avoid the noise of Cyclic Shift Diversity (CSD)\cite{ma2019wifi} and $3$ receiving antennas are attached to each receiver to collect more information. We set the Wi-Fi channel in $5GHz$ frequency band with $20MHz$ bandwidth. The experimental environment is a room about $7m\times8m$, where other existing Wi-Fi networks are operating normally.

The Network Time Protocol (NTP) is used to synchronize the two receivers. We use a camera attached to a receiver to record video footage. We collect CSI at 150Hz and videos at 30Hz so every 5 CSI samples at each receiver are synchronized with one video frame. For more details about the synchronization method of the video footage and CSI samples, please refer to our previous work\cite{guo2019signal,guo2020healthcare}.
\vspace{-0.15cm}
\begin{figure}[!t]
	\centering
	\setlength{\abovecaptionskip}{0.cm}
	\includegraphics[width=0.5\linewidth]{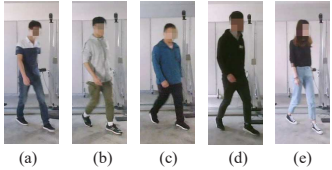}
	\caption{Subjects differ in heights, genders, weights and clothing in our datasets.  }
	\label{volunteers}
	\vspace{-1.0em}
\end{figure}
\begin{table}[!t]
	\setlength{\abovecaptionskip}{0cm}
	\setlength{\belowcaptionskip}{-1cm}  
	\renewcommand{\arraystretch}{1.3}
	\renewcommand{\tabcolsep}{1.0pt} 
	\newcommand{\tabincell}[2]{\begin{tabular}
			{@{}#1@{}}#2\end{tabular}}
	\caption{Feature Extractor Network Implementation} 
	\label{Feature Extractor} \centering 
	\begin{tabular}{p{1.2cm}<{\centering}|p{1.8cm}<{\centering}|p{1.8cm}<{\centering}|p{1.0cm}<{\centering}|p{1.0cm}<{\centering}|p{1.2cm}<{\centering}}
		\hline 
		\tabincell{c}{Feature \\ Extractor} &Input Size  &Output Size   &Kernel   & Stride & \tabincell{c}{Active \\Function}      \\
		\hline
		layer1    &$30\times20\times4$  &$15\times10\times8$ &$3\times3$  &$2\times2$   &ReLU    \\ 
		
		layer2    &$15\times10\times8$  &$15\times10\times8$ &$1\times1$  &$1\times1$   &ReLU    \\
		
		layer3    &$15\times10\times8$  &$8\times5\times32$  &$3\times3$  &$2\times2$   &ReLU    \\
		
		layer4    &$8\times5\times32$   &$8\times5\times32$  &$1\times1$  &$1\times1$   &ReLU    \\
		
		layer5    &$8\times5\times32$   &$4\times3\times128$ &$3\times3$  &$2\times2$   &ReLU    \\
		
		layer6    &$4\times3\times128$  &$4\times3\times128$ &$1\times1$  &$1\times1$   &ReLU    \\
		
		SE$^a$     &$4\times3\times128$  &$4\times3\times128$ & $-$        &      $-$          &   $-$      \\
		\hline 
	\end{tabular}
	\footnotesize{ $^a$SE represents the Squeeze-and-Excitation (SE) block.}
	\vspace{-1.0em}
\end{table}
\begin{table}[!t]
	\setlength{\abovecaptionskip}{0cm}
	\setlength{\belowcaptionskip}{-1cm}   
	\renewcommand{\arraystretch}{1.3}
	\renewcommand{\tabcolsep}{1.0pt}
	\newcommand{\tabincell}[2]{\begin{tabular}
			{@{}#1@{}}#2\end{tabular}} 
	\caption{Generator Network Implementation} 
	\label{Generator} \centering 
	\begin{tabular}{p{1.2cm}<{\centering}|p{1.8cm}<{\centering}|p{1.8cm}<{\centering}|p{1.0cm}<{\centering}|p{1.0cm}<{\centering}|p{1.2cm}<{\centering}} 
		\hline 
		Generator &Input Size       &Output Size       &Kernel       &Stride       & \tabincell{c}{Active \\Function}       \\
		\hline
		FC$^a$        &$4\times3\times128$   &$8\times10\times128$   &      $-$   &     $-$      &ReLU    \\ 
		
		layer1    &$8\times10\times128$  &$15\times20\times64$   &$1\times1$  &$1\times1$    &LReLU    \\
		
		layer2    &$15\times20\times64$  &$15\times20\times64$   &$1\times1$  &$1\times1$    &LReLU    \\
		
		layer3    &$15\times20\times64$  &$30\times40\times32$   &$3\times3$  &$1\times1$    &LReLU    \\
		
		layer4    &$30\times40\times32$  &$30\times40\times32$   &$3\times3$  &$1\times1$    &LReLU    \\
		
		layer5    &$30\times40\times32$  &$60\times80\times8$    &$3\times3$  &$1\times1$    &LReLU    \\
		
		layer6    &$60\times80\times8$   &$60\times80\times8$    &$3\times3$  &$1\times1$    &LReLU        \\
		
		layer7    &$60\times80\times8$   & $120\times160\times1$ &$3\times3$  &$1\times1$    &LReLU        \\
		\hline 
	\end{tabular}
	\footnotesize{ $^a$FC is the fully connected layer.}
	\vspace{-1.0em}
\end{table}
\begin{table}[!t]
	\setlength{\abovecaptionskip}{0cm}
	\setlength{\belowcaptionskip}{-1cm}   
	\renewcommand{\arraystretch}{1.3}
	\renewcommand{\tabcolsep}{1.0pt} 
	\newcommand{\tabincell}[2]{\begin{tabular}
			{@{}#1@{}}#2\end{tabular}}
	\caption{Domain Discriminator Network Implementation} 
	\label{Domain Discriminator} \centering 
	\begin{tabular}{p{1.6cm}<{\centering}|p{1.8cm}<{\centering}|p{1.8cm}<{\centering}|p{1.2cm}<{\centering}} 
		\hline 
		\tabincell{c}{Domain \\ Discriminator} &Input Size       &Output Size      &\tabincell{c}{Active \\Function}      \\
		\hline
		layer1    &$4\times3\times128$  &$1024$    &LReLU    \\ 
		
		layer2    &$1024$               &$1024$    &LReLU    \\
		
		layer3    &$1024$               &$128$     &LReLU    \\
		
		layer4    &$128$                &$4$       &Softmax    \\
		
		\hline 
	\end{tabular}
	\vspace{-1.5em} 
\end{table} 
\subsection{Dataset}
We collect 10 hours of data which contain 5,400,000 CSI samples of each receiver. Note that we collect data on 5 subjects of different heights, genders, weights and clothing in each scenario and only one subject performs continuous poses in the perceptual area at a time, as shown in Figure \ref{volunteers}. In order to evaluate the performance of subject-generalization, we use subject (a) as the target domain and the remaining four subjects (b-e) as the source domains. Specifically, we use 75\% of samples of each subject in the source domain to train the model, and the remaining 25\% of samples in the source domain as well as 25\% of samples in the target domain to test the performance. In addition, we leverage skeleton images on synchronized video footage as annotations of constructed images while we manually generate one-hot vectors as subject labels for CSI images.
\vspace{-0.15cm}
\subsection{Domain-Independent Neural Network}

\subsubsection{Feature Extractor Network}
The feature extractor network uses 6 convolutional layers, followed by a SE block\cite{hu2018squeeze} which is adopted to extract high-level features. Table \ref{Feature Extractor} illustrates the implementation details.

\subsubsection{Generator Network}
As shown in Table \ref{Generator}, a fully connected layer and 7-layer resize convolutions with nearest neighbor interpolation operation are used to convert features into images.

\subsubsection{Domain Discriminator Network}
We utilize 4 fully connected layers to identify domain-related features and perform domain discrimination. In the first three layers, we use Leaky Rectified Linear Unit (LReLU) for recognizing domain-related features better. After the last layer, we adopt the Softmax to calculate the probability distribution of domains. Table \ref{Domain Discriminator} illustrates the implementation details.
\vspace{-0.15cm}
\subsection{Baseline}
Wi-Pose is the state-of-the-art system that can use Wi-Fi signals to construct fine-grained human pose images of the subject who moves in the entire perceptual space. In this paper, we leverage Wi-Pose as our baseline. We note that the training and test data used by the baseline and our system are the same. The baseline and our system are different only in the deep learning model and the training method. More details about Wi-Pose please refer to our previous work\cite{guo2019signal,guo2020healthcare}.
\vspace{-0.15cm}
\subsection{Training Details}
Considering the temporal correlation of human poses, we combine 20 CSI samples which include 5 synchronized samples and 15 preceding samples into one CSI image, corresponding to one video frame and domain label. We utilize TensorFlow \cite{abadi2016tensorflow} to implement the DINN which includes two Adam optimizers. One optimizes the feature extractor and generator networks with an initial $0.001$ learning rate while another one optimizes the discriminator network with an initial $0.0001$ learning rate. $\lambda$ in equation (\ref{loss_L}) is set to $0$ at pre-training stage. The sum of pre-training and adversarial training epochs are $26$. In addition, we adopt the learning rate decay method which multiplies the learning rate by $0.95$ per $5$ epochs.
\begin{figure*}[!t]
	\centering
	\setlength{\abovecaptionskip}{0.cm}
	\includegraphics[width=0.85\linewidth]{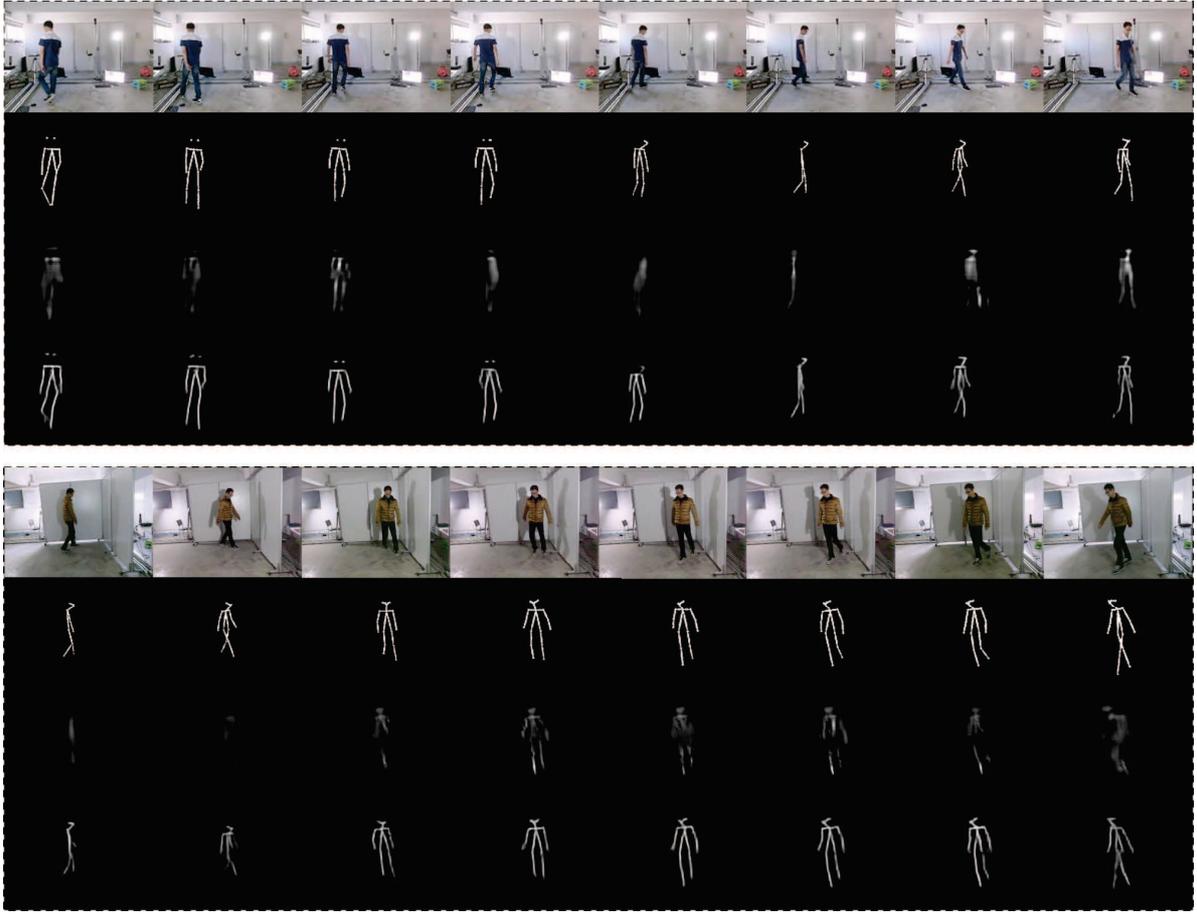}
	\caption{The upper part is in the visible scenario and the bottom part is in the through-wall scenario. In each part, the first pipeline shows the new subject's images recorded by the camera for visual reference here. The second pipeline shows the new subject's skeleton images extracted by OpenPose for the ground truth here. The third and last lines are human pose images constructed by the baseline and our system only using Wi-Fi signals.}
	\label{through_wall}
	\vspace{-1.5em}
\end{figure*} 
\vspace{-0.15cm}
\subsection{Performance}
In this subsection, we evaluate the performance of our system by comparing with the baseline in both the visible and through-wall scenarios. Figure \ref{through_wall} illustrates a test example of a new subject on the baseline and our system in both the visible and through-wall scenarios. In both scenarios, the baseline performs poorly in some positions, and our system significantly improves and constructs fine-grained pose images in these positions. These demonstrate that our system can construct fine-grained human pose images of subjects even if their samples do not undergo any training. 

As mentioned above, both constructed images and annotations are grey-scale maps in our system and the baseline. For quantitatively evaluating the performance, we binarize them in order to simplify the calculation. We convert the non-zero pixels into ones and then calculate the Euclidean distance between constructed images and annotations to measure the difference. Thereby, according to our previous work\cite{guo2019signal,guo2020healthcare}, we leverage Percentage of Correct Skeletons (PCS) to evaluate the performance. It represents the percentage of Euclidean distances less than a certain threshold and is defined as:
\vspace{-1ex}

\begin{shrinkeq}{-1ex}
\begin{small}
	\begin{equation}
	\setlength{\abovedisplayskip}{-0pt}
	\setlength{\belowdisplayskip}{0pt}
    PCS\circ\theta=\frac{1}{S}\sum_{s=1}^{S}\mathbb{I}\left(\parallel{p_{i,j}^s-g_{i,j}^s}\parallel\le\theta\right),
\end{equation}
\end{small}\end{shrinkeq}where $S$ is the number of test frames. $\mathbb{I}$ is a logical operation which outputs $1$ if True and outputs $0$ if False. $p_{i,j}$ and $g_{i,j}$ represent the value for the $(i,j)-th$ pixel of the constructed image and corresponding ground truth, respectively, where $i = 1,2,...,120$ and $j = 1,2,...,160$. $\theta$ refers to the threshold.

According to\cite{guo2019signal,guo2020healthcare}, different thresholds represent different performance of the constructed images. Specifically, $PCS\circ25$ implies that the human pose is accurate, complete, and high-contrast and human position is right in the constructed image. $PCS\circ30$ implies that the human pose is accurate, complete, lower-contrast and human position is right in the constructed image. However, $PCS\circ40$ refers to human position is right but some limbs are a little inaccurate or incomplete or fuzzy in the constructed image. $PCS\circ50$ represents human position is right but more limbs are more inaccurate or more incomplete or fuzzier in the constructed image. We define $PCS\circ30$ as a strict match which means the whole pose is matched and $PCS\circ50$ as a loose match which means only the body is matched. Note that above values are empirical and obtained through experiments. 

Table \ref{table visible} and Table \ref{table through-wall} show the performance in the visible scenario and the through-wall scenario, respectively. Different letters indicate different subjects, and lowercase and uppercase letters indicate subjects are tested by the baseline and our system, respectively. More importantly, a(A) is a new subject who is used to verify the effectiveness of subject-generalization.
\vspace{-0.3cm}
\begin{table*}[!t]
	\setlength{\abovecaptionskip}{0cm}
	\setlength{\belowcaptionskip}{-1cm}   
	\renewcommand{\arraystretch}{1.3} 
	\caption{Results on PCS in the visible scenario} 
	\label{table visible} \centering 
	\begin{tabular}{c|c c|c c c c c c c c|c c} 
		\hline 
		$PCS\circ\theta$&(a)       &\bfseries (A)       &(b)       &\bfseries (B)       &(c)       &\bfseries (C)       &(d)       & \bfseries (D)       &(e)       & \bfseries (E)       & \multicolumn{2}{c}{$Average^a$}  \\
		\hline
		$PCS\circ25$   &3.51\%  &\bfseries 13.49\% &5.62\%  &\bfseries 23.82\% &5.78\%  &\bfseries 22.59\% &6.24\%  & \bfseries 20.66\% &7.95\%  & \bfseries 25.31\% & 5.82\%  &\bfseries 21.17\%   \\ 
		
		$PCS\circ30$   &28.65\% &\bfseries 65.99\% &25.72\% &\bfseries 65.45\% &27.85\% &\bfseries 75.13\% &27.10\% & \bfseries 57.31\% &35.95\% & \bfseries 68.62\% & 29.06\% &\bfseries 66.50\%   \\
		
		$PCS\circ40$   &71.53\% &\bfseries 100\%   &71.38\% &\bfseries 100\%   &77.41\% &\bfseries 100\%   &87.72\% & \bfseries 98.44\% &73.64\% & \bfseries 100\%   & 76.34\% &\bfseries 99.69\%   \\
		
		$PCS\circ50$   &87.25\% &\bfseries 100\%   &90.94\% &\bfseries 100\%   &93.87\% &\bfseries 100\%   &92.98\% & \bfseries 100\%   &90.38\% & \bfseries 100\%   & 91.08\% &\bfseries 100\%     \\
		\hline
		$Average^b$    &39.54   &\bfseries 28.87   &36.63   &\bfseries 28.03   &35.67   &\bfseries 27.76   &37.79   & \bfseries 29.21   &39.41   & \bfseries 27.82   & 37.80   &\bfseries 28.34     \\
		\hline 
	\end{tabular}\\ 
	\footnotesize{$^a$This is the average $PCS\circ\theta$. The left and the right columns belong to the baseline and our system, respectively.}\\
	\footnotesize{$^b$This is the average Euclidean distance of all test samples belong to a subject or all subjects. The penultimate number and the last number are the average of all subjects using the baseline and our system, respectively.}
	\vspace{-1.0em}	
\end{table*}
\begin{table*}[!t]
	\setlength{\abovecaptionskip}{0cm}
	\setlength{\belowcaptionskip}{-1cm}  
	\renewcommand{\arraystretch}{1.3} 
	\caption{Results on PCS in the through-wall scenario} 
	\label{table through-wall} \centering 
	\begin{tabular}{c|c c|c c c c c c c c|c c} 
		\hline 
		$PCS\circ\theta$&(a)      &\bfseries (A)       &(b)       &\bfseries(B)       &(c)       &\bfseries(C)      &(d)       & \bfseries(D)    &(e)       & \bfseries(E)       & \multicolumn{2}{c}{$Average^a$} \\
		\hline
		$PCS\circ25$   &4.92\%  &\bfseries 13.3\%  &7.69\%  &\bfseries 24.35\% &1.36\%  &\bfseries 25.78\% &1.89\%  & \bfseries 13.96\% &1.78\%  & \bfseries 14.03\% & 3.53\% &\bfseries18.28\%      \\ 
		
		$PCS\circ30$   &20.77\% &\bfseries 50.27\% &28.89\% &\bfseries 58.55\% &22.72\% &\bfseries 66.86\% &14.91\% & \bfseries 50.18\% &23.27\% & \bfseries 63.31\% & 22.11\% &\bfseries57.83\%     \\
		
		$PCS\circ40$   &67.58\% &\bfseries 98.18\% &68.11\% &\bfseries 98.71\% &59.81\% &\bfseries 99.61\% &61.51\% & \bfseries 97.0\%  &63.77\% & \bfseries 99.82\% & 64.15\% &\bfseries98.66\%     \\
		
		$PCS\circ50$   &74.86\% &\bfseries 99.82\% &78.80\% &\bfseries 100\%   &70.49\% &\bfseries 100\%   &79.43\% & \bfseries 99.82\% &81.53\% & \bfseries 100\%   & 77.02\% &\bfseries99.93\%     \\
		\hline
		$Average^b$    &46.92 &\bfseries 30.06     &47.46   &\bfseries 28.90   &54.56   &\bfseries 27.93   &48.44   & \bfseries 30.17   &52.70   & \bfseries 28.83   & 50.01   &\bfseries29.18 \\
		\hline 
	\end{tabular}\\
	\footnotesize{$^a$This is the average $PCS\circ\theta$. The left and the right columns belong to the baseline and our system, respectively.}\\
	\footnotesize{$^b$This is the average Euclidean distance of all test samples belong to a subject or all subjects. The penultimate number and the last number are the average of all subjects using the baseline and our system, respectively.}
	\vspace{-1.5em}
\end{table*}
\subsubsection{Overall Performance}
Compared with the baseline, the average percentages of our system significantly improve about 15\% and 37\% in the visible scenario and about 14.7\% and 35.7\% in the through-wall scenario on $PCS\circ25$ and $PCS\circ30$. These imply that our system can construct more fine-grained pose images than the baseline in both scenarios. Our system achieves 66.5\% on average $PCS\circ30$ as well as 100\% on average $PCS\circ50$ in the visible scenario and 57.83\% on average $PCS\circ30$ as well as 99.93\% on average $PCS\circ50$ in the through-wall scenario. These illustrate that 66.5\% and 57.83\% of the constructed pose images strictly match the ground truth in the visible scenario and the through-wall scenario, respectively. Almost all constructed images loosely match the ground truth in both scenarios. In addition, compared with the visible scenario, the overall performance slightly decreases in the through-wall scenario. Because some details are lost when the Wi-Fi signals pass through the wall.
\subsubsection{Subject-generalization Performance}
In our system, 65.99\% and 100\% constructed human pose images of the subject A strictly and loosely match the ground truth in the visible scenario. And 50.27\% and 99.82\% constructed human pose images of the subject A strictly and loosely match the ground truth in the through-wall scenario. These illustrate that our system can construct fine-grained pose images of new subjects in both scenarios, which shows the DINN has high subject-generalization ability. In the two scenarios, $PCS\circ25$ of the subject A are both slightly lower than other subjects' in our system. Because there is no information about him in training samples so that it is difficult to construct accurate, complete and high-contrast human pose images of him. 
\vspace{-0.10cm}
\section{Conclusion}
In this paper, we design the DINN to extract subject-independent features and construct fine-grained human pose images. We also propose a novel training method to train the DINN. It includes two training stages and has no re-training overhead comparing with the domain-adversarial approach. Then, we build a prototype system which can construct fine-grained human pose images of new subjects with commodity Wi-Fi in both the visible and through-wall scenarios. Experimental results show that comparing with the baseline, the average performance of our system increases 37\% and 35.7\% on the strict match in the visible scenario and the through-wall scenario, respectively. 65.99\% (50.27\%) and 100\% (99.82\%) human pose images of the new subject constructed by our system strictly and loosely match the ground truth in the visible (through-wall) scenario. These demonstrate the effectiveness and subject-generalization ability of our model in both scenarios. In the future, we will prove that our model can extract environment-independent features and construct human pose images of new subjects in new environments.
\vspace{-0.10cm}  
\section*{Acknowledgment}
This work is partially supported by National Natural Science Foundation of China under Grant 61671073. It is also partially supported by the Security Risk Perception and Emergency Rescue Decision Support Technology for Large Commercial Complex under grant 2019XFGG23-02.
\vspace{-0.10cm}

\bibliographystyle{IEEEtran}
\bibliography{reference.bib}

\end{document}